\documentclass{article}
\pdfoutput=1

\usepackage[nonatbib, final]{nips_2018}

\usepackage[utf8]{inputenc} 
\usepackage[T1]{fontenc}    
\usepackage{hyperref}       
\usepackage{url}            
\usepackage{booktabs}       
\usepackage{amsfonts}       
\usepackage{nicefrac}       
\usepackage{microtype}      
\usepackage{xcolor}
\usepackage{amsmath}
\usepackage{mathtools}
\usepackage{multicol}
\usepackage{multirow}
\usepackage{tabularx}
\usepackage{graphicx}
\usepackage{epstopdf}
\usepackage{subcaption}
\usepackage{adjustbox}
\usepackage[markup=underlined]{changes}
\graphicspath{ {./Figures/} }

\iffalse 
\definecolor{purple}{rgb}{0.6, 0, 0.6}
\definecolor{orange}{rgb}{1.0, 0.5, 0}
\definecolor{blue}{rgb}{0, 0, 1.0}
\definechangesauthor[color=orange,name=\CHavatar]{CH}
\else
\definecolor{purple}{rgb}{0, 0, 0}
\definecolor{orange}{rgb}{0, 0, 0}
\definecolor{blue}{rgb}{0, 0, 0}
\fi
\newcommand{\raul}[1]{\textcolor{purple}{#1}}
\newcommand{\charles}[1]{\textcolor{orange}{#1}}
\newcommand{\hstz}[1]{\textcolor{blue}{#1}}
\newcommand{\hstzreplace}[2]{\textcolor{blue}{#2}}
\newcommand\Label[1]{&\refstepcounter{equation}(\theequation)&}
\newcommand{\comment}[1]{}

\title{Radiotherapy Target Contouring with Convolutional Gated Graph Neural Network}

\vspace{-.7cm}

\author{
Chun-Hung Chao\textsuperscript{1} \hspace{.5cm} 
Yen-Chi Cheng\textsuperscript{1} \hspace{.5cm} 
Hsien-Tzu Cheng\textsuperscript{1} \hspace{.5cm}
Chi-Wen Huang\textsuperscript{1} \\
\textbf{Tsung-Ying Ho}\textsuperscript{2} \hspace{.5cm} 
\textbf{Chen-Kan Tseng}\textsuperscript{2} \hspace{.5cm} 
\textbf{Le Lu}\textsuperscript{3} \hspace{.5cm}
\textbf{Min Sun}\textsuperscript{1} \\
\textsuperscript{1}National Tsing Hua University \\
\textsuperscript{2}Chang Gung Memorial Hospital \\
\textsuperscript{3}National Institutes of Health Clinical Center\\
\texttt{\{raul.c.chao, charlescheng0117, albertyho\}@gmail.com} \\
\texttt{\{hsientzucheng, kitsune0125\}@gapp.nthu.edu.tw} \\
\texttt{kantseng@adm.cgmh.org.tw, lelu@cs.jhu.edu, sunmin@ee.nthu.edu.tw}
\vspace{-.8cm}
}

\begin{document}
\maketitle

\begin{abstract}
    \vspace{-.3cm}
    \hstz{Tomography medical imaging is essential in the clinical workflow of modern cancer radiotherapy. Radiation oncologists identify cancerous tissues, applying delineation on treatment regions throughout all image slices. This kind of task is often formulated as a volumetric segmentation task by means of 3D convolutional networks with considerable computational cost. Instead, inspired by the treating methodology of considering meaningful information across slices, we used Gated Graph Neural Network to frame this problem more efficiently. More specifically, we propose convolutional recurrent Gated Graph Propagator (GGP) to propagate high-level information through image slices, with learnable adjacency weighted matrix. Furthermore, as physicians often investigate a few specific slices to refine their decision, we model this slice-wise interaction procedure to further improve our segmentation result. This can be set by editing any slice effortlessly as updating predictions of other slices using GGP. To evaluate our method, we collect an Esophageal Cancer Radiotherapy Target Treatment Contouring dataset of 81 patients which includes tomography images with radiotherapy target. On this dataset, our convolutional graph network produces state-of-the-art results and outperforms the baselines. \charles{With the addition of interactive setting, performance is improved even further.}  
    Our method has the potential to be easily applied to diverse kinds of medical tasks with volumetric images. Incorporating both the ability to make a feasible prediction and to consider the human interactive input, the proposed method is suitable for clinical scenarios.} 
    
\end{abstract} \vspace{-.75cm}

\section{Introduction} \vspace{-.3cm}

\hstz{Chemoradiation (combining chemotherapy and radiotherapy) is one of the major regimens of treating cancers. The most significant bottleneck in the treatment flow is the delineation of treatment regions in radiotherapy (RT) plan on patients' tomographic images. This process could take experienced radiation oncologists more than an hour. Since the beginning time of treatment has been shown to affect the outcomes of patients, we argue \charles{that solving this bottleneck is critical for patients}. We focus on the RT treatment of esophageal cancer as our task. Having death rate higher than incidences, less than 20\% of esophageal cancer patients were diagnosed at surgically resectable stage I disease and makes chemoradiation the primary treatment.}


\charles{
\hstzreplace{Specifically, we formulate it as a segmentation problem and utilize deep learning methods to tackle this issue. G}{Fortunately, because of the adequate visual clue in tomography data, the delineation problem can be formulated for the convolutional neural network to solve. Instead of conventional 2D and 3D methods, we propose a novel convolutional gated graph network to tackle this task more efficiently. Specifically,} given a sequence of PET-CT images, we first use an encoder to extract the deep representations of the images. Then we design a graph-based mechanism for our propagator to bridge the communication between these image slices and propagate the information among them. (Throughout this paper, we will use the term ``image'' and ``slice'' interchangeably.) Finally, a decoder will aggregate the high-resolution feature maps from the encoder and the slices propagated by propagator to predict the final segmentation map. The whole process is trained in an end-to-end manner.
}


\raul{We summarize our contributions as below:
    \begin{enumerate}
        \vspace{-.3cm}
        \item We propose to use a graph as the representation for the relationship between each slice in 3D medical images which are characterized by an adjacency matrix learned by our model.\vspace{-.1cm}
        \item We further design a propagator for slices to exchange information in feature space based on the graph representation. Experimental results show that our method significantly outperforms baseline methods.\vspace{-.1cm}
        \item Our work supports interactive setting, by editing one of the slices, the predictions of neighboring slices would be improved according to the user input. Such a setting is more suitable to the clinical scenario and more efficient for doctors to refine the predictions.\vspace{-.1cm}
        \item We collect a new Esophageal Cancer Radiotherapy Target Treatment Contouring dataset. To the best of our knowledge, it is the radiotherapy contouring dataset with the most patients included for esophageal cancer annotated by one or more qualified radiation oncologists.\vspace{-.1cm}
    \end{enumerate} \vspace{-.4cm}
}

\section{Related Work}\vspace{-.3cm}

\noindent\textbf{Neural Network on Graphs.}
\raul{There have been several approaches \charles{which} apply neural networks to graph-structured data. One is to perform graph convolution in the spectral domain, while the other \charles{applies} the neural network to each node in the graph. 
\hstz{Among them, }Acuna et al. \cite{acuna2018efficient} used Gated Graph Neural Network (GGNN) \cite{li2015gated} to refine each vertex in the predicted polygon segmentation mask. Yan et al. \cite{yan2018spatial} used a spatial-temporal graph convolutional networks to process the skeleton graph for action recognition. We exploit the idea in \cite{li2015gated} to our medical image data \charles{by} using \charles{an} adjacency matrix to construct the graph, through which \charles{the} information \charles{is} propagated between image slices.}\\
\noindent\textbf{3D Biomedical Image Segmentation.}
\raul{In clinical practice, especially oncology, physicians heavily rely on 3D medical images to make diagnoses or treatment plan\charles{s}. Various CNN based image segmentation methods have been developed to give direct or indirect assistance in the medical procedures and related biological research\charles{es}. Havaei et al. \cite{HAVAEI201718} proposed a model which input is 2D Image patch from 3D brain MRI volume to segment the region of brain tumors. \cite{3DU-Net, VNet} used 3D convolutional kernels for the volumetric segmentation of Xenopus kidney and prostates respectively. \cite{Chen2016CombiningFC, tseng2017joint} leveraged the combination of CNN and the recurrent unit for exploiting the intra- and inter-slice contexts of image volumes of 3D neuron structures and the human brain. In our work, we design an information propagator to exchange information between slices in the 3D medical image volume, which is demonstrated to predict RT planning contours of good quality.}\\
\noindent\textbf{Interactive Segmentation.}
\raul{To generate an accurate treatment plan which meets radiologists' knowledge and experience, cooperation with experts is one of the most crucial issues in our application. In fact, user-assisted segmentation for the 3D medical image has been studied for years. Methods include modified GrowCut algorithm \cite{zhu2014effective}, an interactive network \cite{amrehn2017ui}, and weighted loss function which can incorporate with optional user scribble input \cite{wang2018interactive}. While the previous methods require the user to edit each predicted image, our model aims to learn essential information in high dimensional feature space and \charles{then} propagated \charles{those features} for prediction refinement of neighboring slices.}\vspace{-.2cm}


\begin{figure}[tbp]
\begin{center}
\includegraphics[width=.9\linewidth]{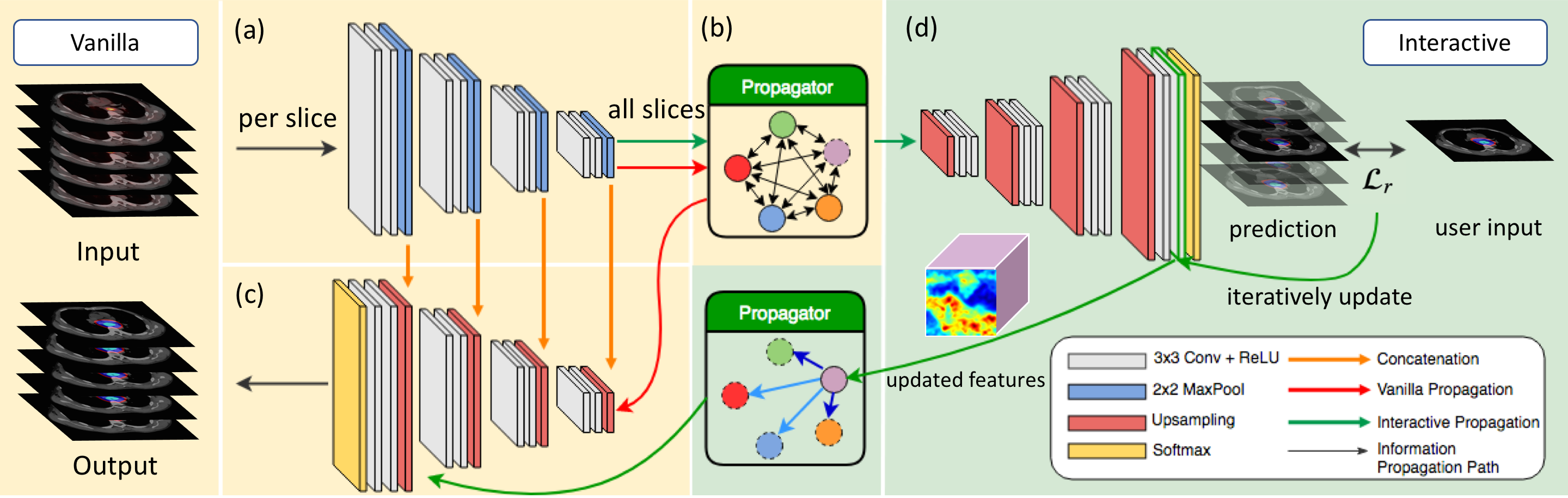}
\vspace{-0.25cm}
\end{center}
\caption{
System overview. 
}\vspace{-0.25cm}
\end{figure}\label{fig.system}
\vspace{-0.25cm}
\section{Approach}\label{sec.appr}\vspace{-0.3cm}

\subsection*{Notations}\label{sec.notation}\vspace{-0.35cm}
\charles{Given \raul{n} stacked \raul{PET-CT images} $I \in \mathbb{R}^{2 \times h \times w}$ with height $h$ and width $w$, we \raul{compile} them to form a sequence of slices $s \in \mathbb{R}^{\raul{n} \times 2 \times h \times w} $. \raul{We then feed the sequence of stacked images $s$ to our model to produce $Y$, the treatment region prediction of the sequence's middle image. Since our model is an encoder-decoder architecture, \hstzreplace{which will be talked about later}{mentioned later in Sec. \ref{sec.model}}, we are able to obtain a deep representation $f_v \in \mathbb{R}^{c^{\prime} \times h^{\prime} \times w^{\prime}}$ for slice $v$ in the sequence and $F \in \mathbb{R} ^ {|V| \times c^{\prime} \times h^{\prime} \times w^{\prime}}$ for the whole sequence. In the following sections, the convolution operation will be denoted as $\otimes$ and could be followed with a bias term $b$ and an activation function $\sigma$. We also defined a function $\tau$ to reshape its high dimensional inputs to a matrix, $\tau: \mathbb{R}^{c \times h \times w} \to \mathbb{R}^{(c \times h \times w) \times 1}$ \charles{, and its inverse mapping, } $\tau^{-1}: \mathbb{R}^{1 \times (c \times h \times w)} \to \mathbb{R}^{c \times h \times w}$}.}
\vspace{-0.4cm}
\subsection{Model Architecture}\label{sec.model}
\vspace{-0.25cm}
\noindent\textbf{Overview.}
\charles{For each sequence $s$, our model first uses U-Net-like contraction part to obtain the feature maps of the images with various resolutions (see Figure \ref{fig.system} (a)). Then our Gated Graph Propagator (GGP, \hstzreplace{we will explain it in detail in the next section.}{elaborated in the next paragraph}) propagates \hstzreplace{the information between the sequence of slices to provide region information among them.}{spatial information throughout all slices in the sequence. (see Figure \ref{fig.system} (b))} Subsequently, the decoder combines deep representations of slices and high-resolution features from the skip connection of our encoder to perform up-sampling (see Figure \ref{fig.system} (c)). Finally, a pixel-wise softmax will output the final segmentation mask $M \in \mathbb{R}^{h \times w}$ as our prediction. We use DICE loss to train our network.} \\
\noindent\textbf{Gated Graph Propagator.}
\charles{We propose to use GGNN as the backbone of our GGP.} 
Given \raul{$F$} produced from our encoder, we set up a learnable adjacency matrix \smash{$A \in \mathbb{R}^{|V| \times |V||E| }$}, \raul{where $V$ is the set of indices for each slice in the sequence and $E$ is the set of indices for different types of edges. $A$ can be used to establish and weight the relationship between slices.} \raul{For each pair of slice $u$ and $v$, there will be a $e \in E$ type of convolutional kernel $k^{e}_{uv}$ to extract essential information from $f_u$, weighted by $A$:}
\raul{
    \small
    \vspace{-0.1cm}
    \begin{align}
        g_{(|V|e + u)(v)} =  \sigma(k^{e}_{uv} \otimes f_u + b^{e}_{uv})
    \end{align}
    \vspace{-0.3cm}
    \begin{equation}
            h^{t}_{v} = \tau^{-1}(a_{v, *} \begin{bmatrix} 
                                            \tau(g^{(t-1)}_{(0)(v)})&
                                            \tau(g^{(t-1)}_{(1)(v)})&
                                            \dots&
                                            \tau(g^{(t-1)}_{(|V||E|)(v)})
                                        \end{bmatrix} ^{\top})
    \end{equation}
    \normalsize
}
\raul{Same as \cite{li2015gated}, we use Gated Recurrent Unit like mechanism to update $F$:}
\charles{
\noindent\begin{minipage}[t]{.5\textwidth}
    \vspace{-0.4cm}
    \small
    \begin{align}
        z^{t}_{v} &= \sigma(W^z \otimes h^{(t)\top}_{v} + U^z \otimes f^{t-1}_{v}) \label{eq1} \\
        r^{t}_{v} &= \sigma(W^r \otimes h^{(t)\top}_{v} + U^r \otimes f^{t-1}_{v}) \label{eq3}
    \end{align}
    \normalsize
\end{minipage}%
\begin{minipage}[t]{.5\textwidth}
    \vspace{-0.4cm}
    \small
    \begin{align}
        \tilde{f^{t}_{v}} &= \sigma(W \otimes h^{(t)\top}_{v} + U(r^t_v \odot f^{t-1}_{v})) \label{eq2}\\
        f^{t}_{v} &= (1-z^{t}_{v}) \odot f^{t-1}_{v}  + z^{t}_{v} \odot \tilde{f^{t}_{v}} \label{eq4}
    \end{align} 
    \normalsize
\end{minipage}
}


\noindent\textbf{Interactive Setting.} \raul{Furthermore, we collect representations of a sequence of slices, $Z$, with one of the representations for slice $v$ was replaced with the user input representation $\hat{z_v}$ to form the propagator for the interactive setting. We adopt the features from the last convolutional layer as the representations (see Figure \ref{fig.system} (d)). However, $\hat{z_v}$ can't be obtained directly, we approximate the representation for user input, $\tilde{z_v}$, by iteratively updating the original features from the last convolutional layer $z_v$ until the prediction is close enough to the user input while inferencing $z^{t}_{v}$:}

\raul{
    \vspace{-.5cm}
    \small
    \begin{equation}
        z^{t+1}_{v} = z^{t}_{v} - \alpha (\dfrac{\partial \mathcal{L}_r(\phi(z^{t}_{v}),\hat{Y_v})}{\partial z^{t}_{v}})
    \end{equation}
    \normalsize
}
\charles{where $\alpha$ denotes the learning rate, $\hat{Y_v}$ is the user input, $\phi$ is the last convolutional layer and softmax and $\mathcal{L}_r$ is the NLL loss. }\raul{Once $\tilde{z_v}$ is obtained, we then train with a handcrafted adjacency matrix, where values of each entry are assigned according to its distance to slice $v$.} \raul{Then other slices in the sequence would be able to update its own representation regarding the information from the $\tilde{z_v}$.}\vspace{-.4cm}
\section{Experiments}
\begin{figure*}[h!]
    \centering
    \begin{subfigure}[b]{0.45\textwidth}
        \centering
        \includegraphics[width=\textwidth]{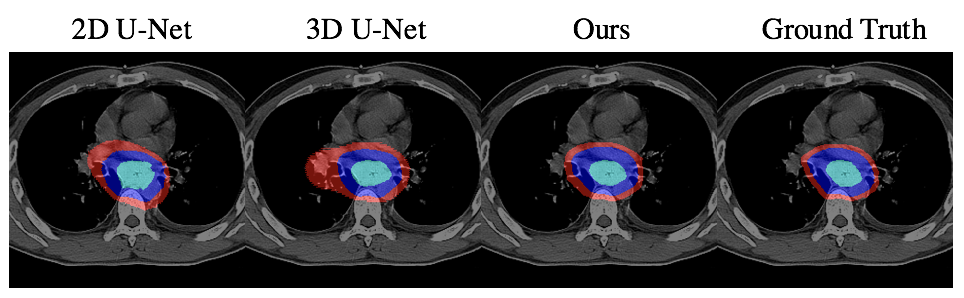}\vspace{-0.1cm}
        \caption[]%
        {{\small Patient 1}}\vspace{-.1cm}
        \label{fig:mean and std of net14}
    \end{subfigure}
    \begin{subfigure}[b]{0.45\textwidth}  
        \centering 
        \includegraphics[width=\textwidth]{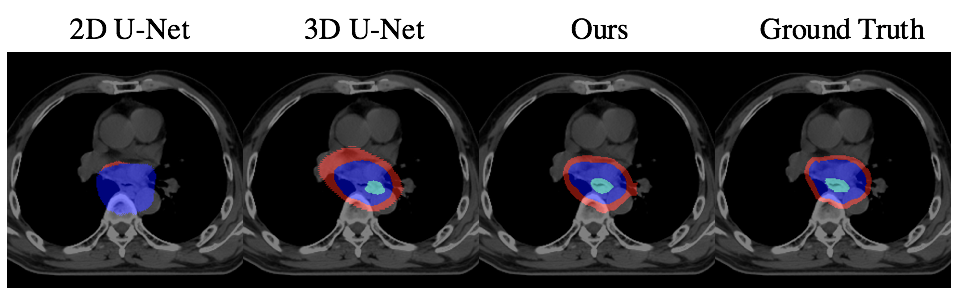}\vspace{-0.1cm}
        \caption[]%
        {{\small Patient 2}}\vspace{-.1cm}
        \label{fig:mean and std of net24}
    \end{subfigure}
    \vspace{-.25cm}
    \begin{subfigure}[b]{0.45\textwidth}
        \centering 
        \includegraphics[width=\textwidth]{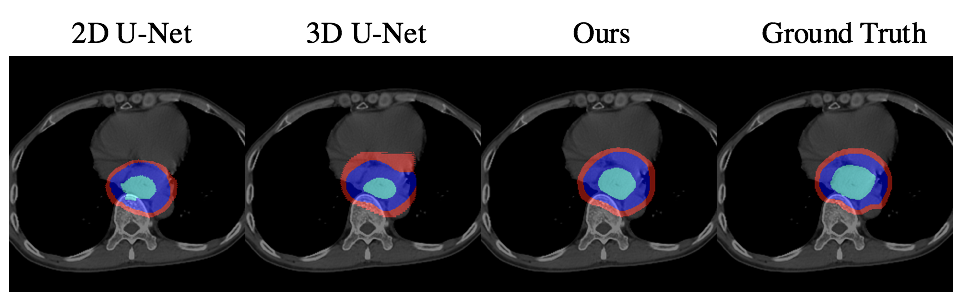}\vspace{-0.1cm}
        \caption[]%
        {{\small Patient 3}}
        \label{fig:mean and std of net34}
    \end{subfigure}
    \begin{subfigure}[b]{0.45\textwidth}   
        \centering 
        \includegraphics[width=\textwidth]{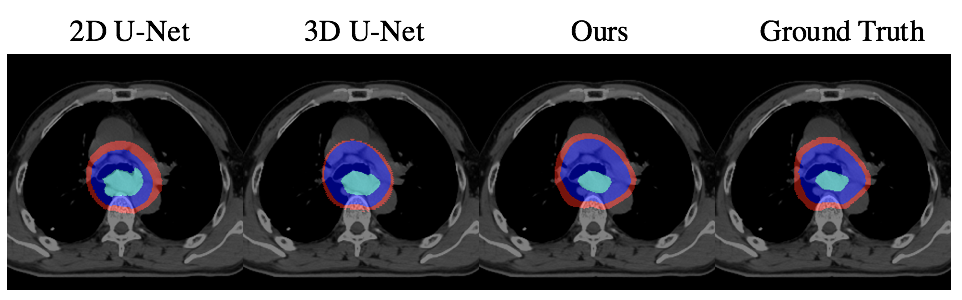}
        \caption[]%
        {{\small Patient 4}}\vspace{-.05cm}
        \label{fig:mean and std of net44}
    \end{subfigure}
    \caption[Qualitative results.]
    {\small Qualitative results. Cyan: GTV; Blue: CTV; Red: PTV.} \vspace{-.2cm}
    \label{fig.qualitativeresult}
\end{figure*}

\vspace{-.35cm}
\noindent\textbf{Dataset.}
\hstz{Our Esophageal Cancer Radiotherapy Target Treatment Contouring dataset, which contains PET-CT/RT-CT images of 81 patients treated between 2015 and 2017, was collected through the cooperation with Chang Gung Memorial Hospital Linkou branch. The scan resolution is $512\times512$ pixels with slice thickness ranging from 2.5 to 5 mm. PET and CT are considered two kinds of input modalities; the segmented targets for RT are Gross Tumor Volume (GTV), Clinical Target Volume (CTV), and Planning Target Volume (PTV).}
\\
\noindent\textbf{Experimental Settings and Results.}
\charles{To evaluate our method, we chose the most commonly used CNN-based segmentation frameworks: 2D U-Net\cite{U-Net} and 3D U-Net\cite{3DU-Net} as our baselines\hstzreplace{ to demonstrate the effectiveness of our model}{}. \charles{Since we focus on predicting the accurate radiotherapy target contouring, we trained and tested with valid slices that contain tumors. That is, given the volume where oncologists want to deliver radiation, we are able to segment the GTV, CTV and PTV regions precisely. }\hstzreplace{As for the metrics, we used the score for qualitative comparisons. Table 1 showed the results of the baseline models and our methods.}{For quantitative comparison, following metrics in \cite{HAVAEI201718}, we use DSC, Sensitivity, and Specificity and ran with five-fold cross-validation. Table. \ref{tab.result_ecd} and Figure. \ref{fig.qualitativeresult} shows that our non-interactive method outperforms all the baselines both qualitatively and quantitatively.} In the interactive setting, the results are evaluated on the selected sequences with ``min median DSC'' of GTV for each patient, in the exclusion of the slice we have reconstructed the features. As shown in Table. \ref{tab.result_int}, the results are further improved after the interaction.}

\begin{table}[h!]
  \centering
  \vspace{-.3cm}
  \caption{Quantitative result on Esophageal Cancer Dataset.}
  \setlength{\tabcolsep}{6pt}
  \begin{adjustbox}{width=.98\textwidth}
  \begin{tabular}{*{11}{l}}
    \toprule
    & \multicolumn{3}{c}{DSC (\%)} & \multicolumn{3}{c}{Sensitivity (\%)} & \multicolumn{3}{c}{Specificity (\%)}  \\
    \cmidrule(lr){2-4}
    \cmidrule(lr){5-7}
    \cmidrule(lr){8-10}
    Metrics & GTV & CTV & PTV & GTV & CTV & PTV & GTV & CTV & PTV &\\
    \midrule
    2D U-Net & 73.3$\pm$2.3& 80.7$\pm$1.1& 86.1$\pm$1.5 & 79.2$\pm$2.5 & 79.8$\pm$4.6 & 83.8$\pm$3.4 & 99.7$\pm$0.0 & 99.3$\pm$0.3 & 99.3$\pm$0.1\\
    3D U-Net & 79.2$\pm$2.8& 85.7$\pm$2.0& 89.3$\pm$1.6 & 78.0$\pm$3.2 & 84.4$\pm$3.8 & 86.9$\pm$2.2 & 99.8$\pm$0.0 & \textbf{99.4}$\pm$0.1 & \textbf{99.4}$\pm$0.1 \\
    Ours     & \textbf{82.7}$\pm$3.2 & \textbf{86.0}$\pm$2.0 & \textbf{90.3}$\pm$1.3 & \textbf{80.5}$\pm$2.5 & \textbf{85.7}$\pm$3.4 & \textbf{92.2}$\pm$1.6 & \textbf{99.8}$\pm$0.1 & 99.3$\pm$0.1 & 98.9$\pm$0.4 \\
    \bottomrule
  \end{tabular}
  \end{adjustbox}
  \label{tab.result_ecd}
  \vspace{-.35cm}
\end{table}
\begin{table}[h!]
  \centering
  \caption{Quantitative result of interactive setting.}
  \setlength{\tabcolsep}{6pt}
  \begin{adjustbox}{width=.45\textwidth}
  \begin{tabular}{*{11}{l}}
    \toprule
    Metrics & DSC (\%)& Sensitivity (\%)& Specificity (\%)\\
    \midrule
    Non-interactive  & 71.06$\pm$6.7 & 72.77$\pm$6.6     & 99.81$\pm$0.0 \\
    Interactive & \textbf{75.01}$\pm$6.4 & \textbf{76.51}$\pm$7.3 & \textbf{99.83}$\pm$0.0 \\
    \bottomrule
  \end{tabular}
  \end{adjustbox}
  \label{tab.result_int}
  \vspace{-.5cm}
\end{table}

\vspace{-.4cm}

\section{Conclusions}\vspace{-.4cm}
In this paper, we propose an end-to-end network for radiotherapy target contouring. We combine an encoder-decoder with a convolutional Gated Graph Propagator, featured by a learnable adjacency matrix to exchange information with neighbor slices. In the interactive setting, we apply an iterative method for user input based feature reconstruction of a certain slice further enhancing the segmentation results of its neighbors. The experimental result on our Esophageal Cancer Radiotherapy Target Treatment Contouring dataset shows our system's capability of predicting RT planning contours interactively and efficiently, thus its suitability for clinical scenarios.\vspace{-.4cm}




\comment{
{\small
    \bibliographystyle{ieee}
    \bibliography{egbib}
}
}

{\small
    \bibliographystyle{ieee}
    \bibliography{egbib}
}

\end{document}